\documentclass[letterpaper, 10 pt, journal, twoside]{IEEEtran}  

\usepackage{times}

\usepackage{multicol}

\usepackage{amsmath,amssymb,amsfonts} 
\usepackage{algorithm}               
\usepackage{algpseudocode}           
\usepackage{graphicx}                
\usepackage{capt-of}                 
\usepackage{multirow}
\usepackage{booktabs} 
\usepackage{xcolor}
\usepackage{tabularx}
\usepackage{array}
\usepackage[normalem]{ulem}

\usepackage{cite}

\usepackage[hidelinks]{hyperref}

\title{SyncTwin: Fast Digital Twin Construction and Synchronization\\ for Safe Robotic Manipulation}

\author{Ruopeng Huang$^{1}$, Boyu Yang$^{2}$, Wenlong Gui$^{2}$, Jeremy Morgan$^{2}$, Erdem B{\i}y{\i}k$^{2}$, and Jiachen Li$^{3,\dagger}$%
\thanks{Manuscript received: June 24, 2026; Revised: September 18, 2026; Accepted: October 19, 2026.}%
\thanks{This paper was recommended for publication by Editor Markus Vincze upon evaluation of the Associate Editor and Reviewers' comments.
($^{\dagger}$Corresponding author: Jiachen Li.)}%
\thanks{$^{1}$Ruopeng Huang is with the University of California, Los Angeles, Los Angeles, CA 90095, USA {\tt\footnotesize ruopengh@ucla.edu}. This work was done during an internship at the Trustworthy Autonomous Systems Laboratory.}%
\thanks{$^{2}$Boyu Yang, Wenlong Gui, Jeremy Morgan, and Erdem B{\i}y{\i}k are with the Viterbi School of Engineering, University of Southern California, Los Angeles, CA 90089, USA.}%
\thanks{$^{3}$Jiachen Li is with the Georgia Institute of Technology, Atlanta, GA 30332, USA {\tt\footnotesize jiachen\_li@gatech.edu}.}%
\thanks{Supplementary videos and implementation code are available on our project website: \url{https://sync-twin.github.io/}.}%
\thanks{Digital Object Identifier (DOI): see top of this page.}%
}

\begin{document}

\bstctlcite{IEEEexample:BSTcontrol}

\IEEEaftertitletext{%
    \vspace{-2.5\baselineskip}
    \begin{center}
    \phantomsection
    \includegraphics[width=0.8\textwidth]{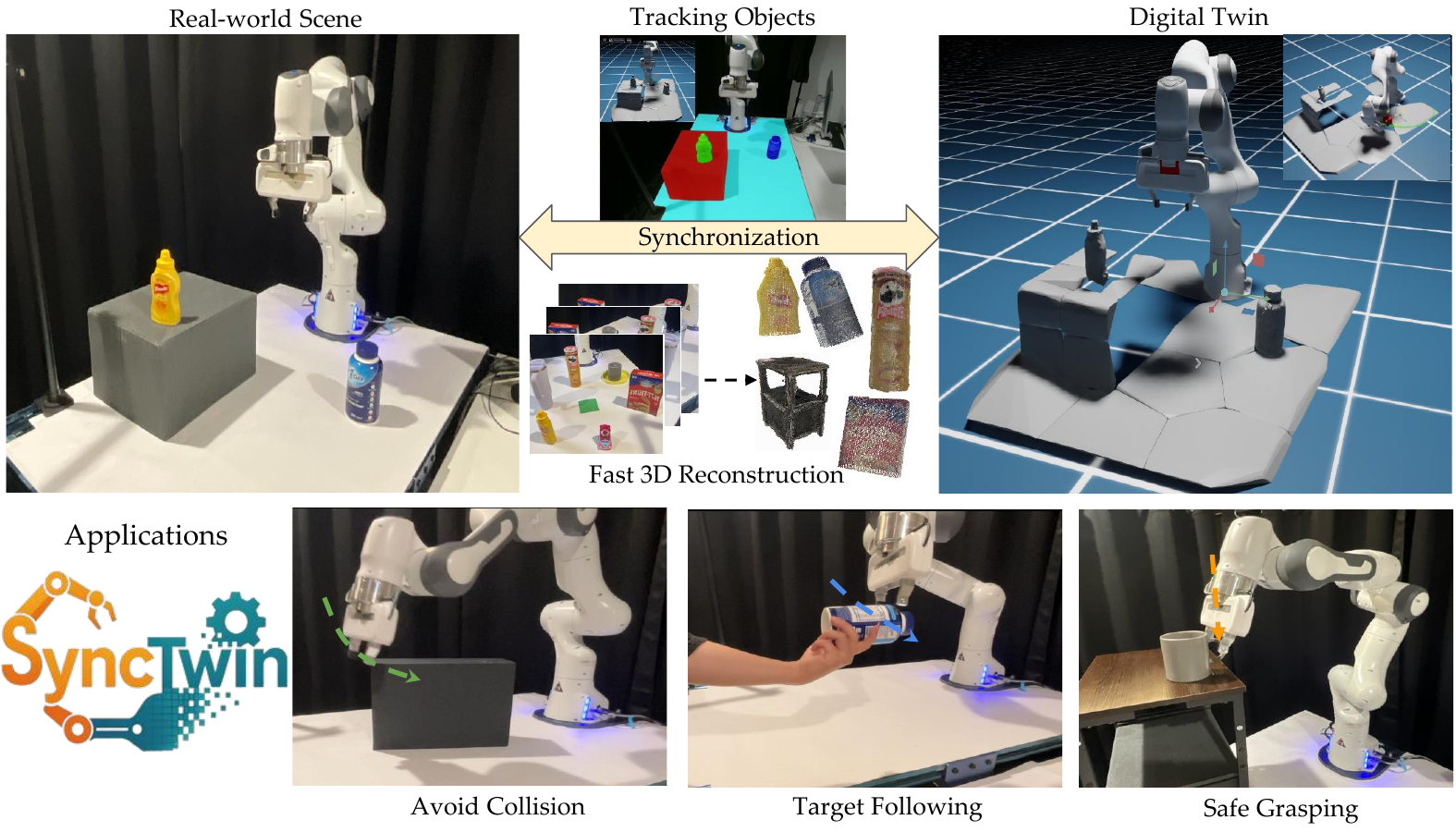}
    \captionof{figure}{SyncTwin is a system that enables fast digital twin construction and real-time synchronization through efficient 3D scene reconstruction and object tracking. By bridging motion planning between simulation and the real world, the system supports dynamic obstacle avoidance, target following, and safe grasping under single-view occlusion in the real world.
    }
    \label{fig:overview}
    \end{center}
    \vspace{-0.1\baselineskip}
}

\maketitle

\begin{abstract}

Accurate and safe robotic manipulation under dynamic and visually occluded conditions remains a core challenge in real-world deployment. We introduce SyncTwin, a novel digital twin framework that unifies fast 3D scene reconstruction and real-to-sim synchronization for robust and safety-aware robotic manipulation in such environments. In the offline stage, we employ VGGT to rapidly reconstruct object-level 3D assets from RGB images, forming a reusable geometry library. During execution, SyncTwin continuously synchronizes the digital twin by tracking real-world object states and uses stored assets to complete partially observed geometry. The synchronized twin enables motion planners to compute collision-aware robot trajectories in simulation, with execution on the real robot closing the real-to-sim-to-real loop. Experiments in dynamic and occluded scenes show that SyncTwin improves manipulation performance and motion safety, demonstrating the effectiveness of digital twin synchronization for real-world robotic execution.

\end{abstract}

\begin{IEEEkeywords}
 Perception for Grasping and Manipulation, Engineering for Robotic Systems,  Digital Twin.
\end{IEEEkeywords}

\section{Introduction}

\IEEEPARstart{A}{chieving} accurate and safe robotic manipulation in dynamic real-world environments remains a long-standing challenge due to incomplete perception and scene dynamics~\cite{liu2026pacer}. 
Without an accurate understanding of their surroundings, robots risk colliding with hardware or even endangering humans.
Thus, both in real-robot online training~\cite{gong2026hierarchical} and in real-world robot model deployment~\cite{gao2026guides}, enabling robots to safely plan and execute motions under dynamic scene changes is a critical concern.
Unlike simulation, where complete scene geometry and object states are fully accessible, real-world perception often provides only partial and occluded observations.
Robust motion planning in real-world environments still remains a challenging open problem.

Several recent efforts have attempted to bridge the gap between perception and control.
Voxel-based mapping systems like NVBlox~\cite{millane2024nvblox} enable online obstacle reconstruction, yet they only provide a voxel grid, which is a representation too coarse for reliable manipulation. 
End-to-end reactive policies like DRP~\cite{yang2025deep} handle dynamics through continuous control but require robot-specific retraining and lack generalization across new hardware. 
And both share a critical limitation: they operate without a consistent and complete model of the scene, leading to unsafe or unreliable executions. 

Bridging the sim-to-real gap, a robot needs a perception model that can efficiently perceive what objects exist in the environment and also track how the real-world scene evolves over time.
In other words, instead of planning in a static or outdated simulation, the robot should plan within a dynamic digital twin that mirrors the physical world in real time, where perception and control are tightly coupled through continuous real-to-sim synchronization.
However, real-world perception is inherently partial single-view occlusions often reveal only fragments of object geometry, making grasping and motion planning unreliable. 
Inspired by SAM4D~\cite{xu2025sam4d}, which maintains a memory bank of object assets, we incorporate the idea of leveraging object-level memories to complete partial observations at execution time. 
We develop \textbf{SyncTwin}, a digital twin framework equipped with an object memory bank that performs real-time object tracking, injects accurate poses and geometries into simulation, and closes the loop by executing planned trajectories back on the real robot, addressing a key limitation of prior digital twin systems, which do not support continuous real-to-sim-to-real synchronization (see Fig.~\ref{fig:overview}).

SyncTwin operates in two stages.
In Stage~I, VGGT~\cite{wang2025vggt} generates scene-level point clouds from RGB inputs. Object-level point clouds are extracted via projection, segmentation, and denoising, then converted into lightweight meshes and stored in a memory bank.
This design addresses a practical gap: learning-based reconstruction can be fast, but its estimated camera poses and scales are often imperfect, making direct object-level asset extraction unreliable.
In Stage~II, online object tracking and pose estimation~\cite{park2017colored,wen2024foundationpose} continuously synchronize the digital twin in Isaac Sim~\cite{mittal2023orbit}, enabling collision-aware planning with cuRobo~\cite{sundaralingam2023curobo}.
By using robot models rather than robot-specific policies, the framework is designed to support different manipulators without retraining.

The main contributions of this paper are as follows: 
\begin{itemize}
    \item 
    We present the first framework with continuous synchronization and asset completion for manipulation, enabling collision-aware planning and a closed real-to-sim-to-real loop for dynamic, partially observed scenes.
    \item We introduce a fast, low-cost RGB-only method that constructs 3D geometry assets using learning-based geometry estimation and projection-based segmentation.
    \item We develop a real-time 3D segmentation and tracking module that processes streaming RGB-D camera data.
    \item Our experiments demonstrate that the proposed system improves grasp accuracy and safety in single-view and occluded settings, while also achieving state-of-the-art time efficiency in 3D geometry asset extraction.
\end{itemize}

\begin{figure*}[!t]
    \centering
    \includegraphics[width=\linewidth]{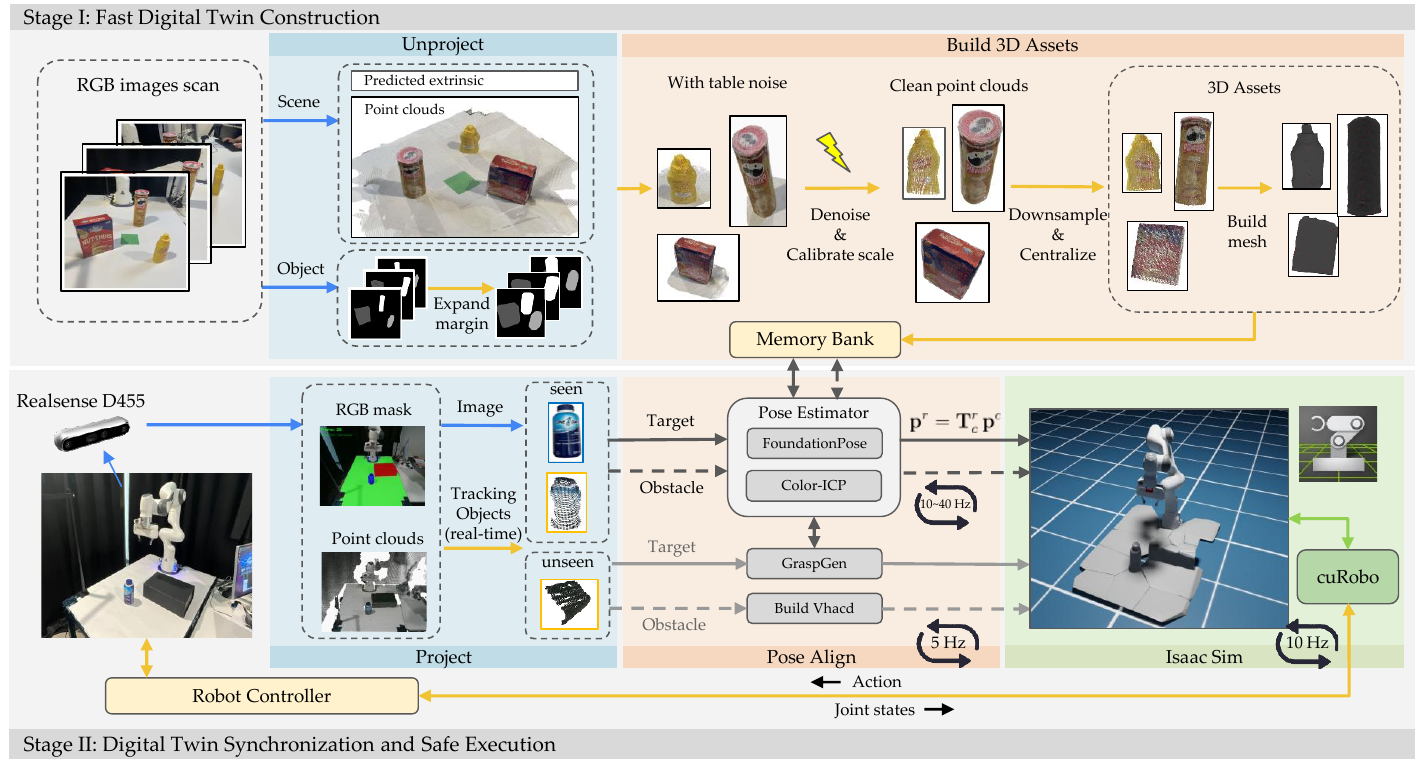}
    \caption{\textbf{Framework of the SyncTwin.} 
    Stage~I constructs simulation-ready 3D assets from RGB images via VGGT. Multi-view masks are unprojected into point clouds, then denoised, scaled, and meshed into clean object assets for memory bank storage. 
   Stage II performs real-time object segmentation, pose tracking, and asset-based completion, enabling grasp generation and reactive motion planning in a closed real2sim2real loop. By continuously updating the digital twin and leveraging simulation for decision making, the system ensures collision-aware execution under dynamic and partially occluded environments.}
    \label{fig:structure}
\end{figure*}

\section{Related Work}
\label{sec:related}

\textbf{3D Scene Reconstruction and Segmentation.}
Traditional 3D scene reconstruction relies on either RGB-D inputs or optimization-based RGB-only pipelines~\cite{mildenhall2021nerf,kerbl20233d}. Though capable of producing visually plausible geometry, these methods incur high computational costs for complete reconstruction.
Object-centric 3D segmentation is also critical for robotic applications. Prior work has explored segmentation directly on point clouds via learned clustering and aggregation~\cite{zhou2024point}, or adopted multi-view projection of 2D masks for 3D instance segmentation~\cite{zhang2023sam3d}. 
However, projection-based methods rely on precise camera extrinsics~\cite{xu2025sampro3d}, and few studies have evaluated their performance on point clouds from learning-based reconstruction.
Our work fills this gap by unifying learned 3D reconstruction and mask-guided multi-view segmentation for efficient object asset extraction, providing geometry-aware perception for safe and reliable motion planning.

\noindent\textbf{Digital Twin for Robotic Manipulation.}
Digital twins are increasingly used for robot data generation and policy learning~\cite{mu2025robotwin, katara2024gen2sim}, but many rely on scene representations constructed before deployment~\cite{ning2025prompting}. 
Meanwhile, real-time 6D tracking method ~\cite{wen2024foundationpose} estimate object poses. 
Recent work such as Real-is-Sim explores continuously synchronized simulators for policy execution~\cite{abouchakra2026realissim}. 
In contrast, SyncTwin continuously synchronizes object pose and geometry from online perception and updates the simulation for collision-aware planning.

\noindent\textbf{Safe Motion Planning for Manipulation.}
Safe motion planning remains a critical challenge for robotic manipulation in unstructured environments.
Classical planners~\cite{kalakrishnan2011stomp, schulman2013finding} 
rely on fully observable simulated environments, yet struggle with partial real-world observations.
To narrow this gap, several methods construct static point cloud maps and import them into simulation for offline planning~\cite{brunke2025semantically}, but lack real-time adaptability.
Alternative strategies ensure safety via latent-space visual planning~\cite{NakamuraK-RSS-25, bahety2025safemimic}, force-based control~\cite{shan2026learning, kang2025robotic}, voxelizing scene geometry~\cite{millane2024nvblox}, uncertainty-aware planning~\cite{dai2019chance,de2025scenario}, or distilling planning policies from point cloud observations~\cite{yang2025deep}.
In contrast, we maintain a synchronized digital twin for safe execution in dynamic scenes.

\section{Method}
\label{sec:method}
SyncTwin consists of two stages: (1) fast digital twin construction and (2) digital twin synchronization. An overall framework of the proposed
method is provided in Fig.~\ref{fig:structure}.

\subsection{Problem Formulation}
\label{sec:problem}

\noindent\textbf{Stage~I: Fast Digital Twin Construction.}
Given a small set of RGB images $\{\mathbf{I}_i\}_{i=1}^{N}$, intrinsics $K$ and estimated extrinsics $\{\mathbf{T}_i^\text{world}\}_{i=1}^{N}$. Our goal is to generate object-level, simulation-ready 3D assets $\mathbb{B}=\{\mathcal{X}_j, \mathcal{M}_j^{3D}\}$ from these images, and balance reconstruction efficiency and asset quality.

\noindent\textbf{Stage~II: Online Digital Twin Synchronization.}
During execution, the fixed calibrated camera receives streaming RGB-D frames and corresponding partial point clouds $\mathcal{X}_p$.  
Object poses $\mathbf{T}_j^\text{world}$ are updated using either colored-ICP~\cite{park2017colored} to align partial point clouds $\mathcal{X}_p$ with complete asset point clouds $\mathcal{X}_m$, or FoundationPose~\cite{wen2024foundationpose} with RGB images $\mathbf{I}_i$ and asset meshes $\mathcal{M}_j^{3D}$.
This synchronized scene is streamed into Isaac Sim~\cite{mittal2023orbit},  
where cuRobo’s MPC planner~\cite{sundaralingam2023curobo} produces collision-free trajectories  
$\mathbf{A}_{t:t+H} = \{\mathbf{a}_0,\ldots,\mathbf{a}_H\}$. 
The key challenge is robustly tracking objects under occlusion and partial observation, while ensuring that the object poses and geometries can be accurately updated into the simulator to enable safe real-to-sim-to-real planning and execution.

\subsection{Fast Digital Twin Construction}

The first stage aims to rapidly reconstruct the 3D environment and extract object-level representations suitable for real-time simulation.
Our goal is not to propose a new reconstruction or rendering algorithm, but to bridge fast scene-level RGB reconstruction and manipulation-ready object-level digital assets for robotic planning.
We adopt VGGT~\cite{wang2025vggt} over other 3D reconstruction methods, because it provides a practical trade-off among speed, flexibility, cost, and sufficient geometric fidelity for our deployment setting. 

Nevertheless, direct use of VGGT outputs is insufficient for robotic manipulation.
First, the camera extrinsics estimated by VGGT are often inaccurate, which leads to projection misalignment during mask-based segmentation.
Second, the generated point cloud is not in a true metric world scale, causing imported assets to appear incorrectly sized relative to the robot in simulation.
Therefore, Stage~I focuses on transforming fast scene-level outputs into simulation-ready object assets by addressing extrinsic inaccuracies, scale consistency, support-plane noise, and mesh complexity. An operator moves a handheld camera around the objects to capture RGB or RGB-D views. SAM2 generates segmentation masks from either manually clicked point prompts or bounding-box prompts returned by an object detector~\cite{cheng2024yolo}.

\noindent\textbf{Mask Projection Expansion.}
To mitigate inaccuracies in VGGT-estimated camera extrinsics, each 2D segmentation mask $\mathcal{S}_i$ is spatially expanded before projection, which ensures full coverage of object boundaries and prevents missing edge regions.
While mask expansion compensates for projection drift, it also introduces floating outliers (e.g., background points around the object) and merged support-plane regions (e.g., table surfaces).  
We thus apply our point clouds denoising mechanism to isolate the true object shape.

\begin{figure}[!t]
    \centering
    \includegraphics[width=1.0\linewidth]{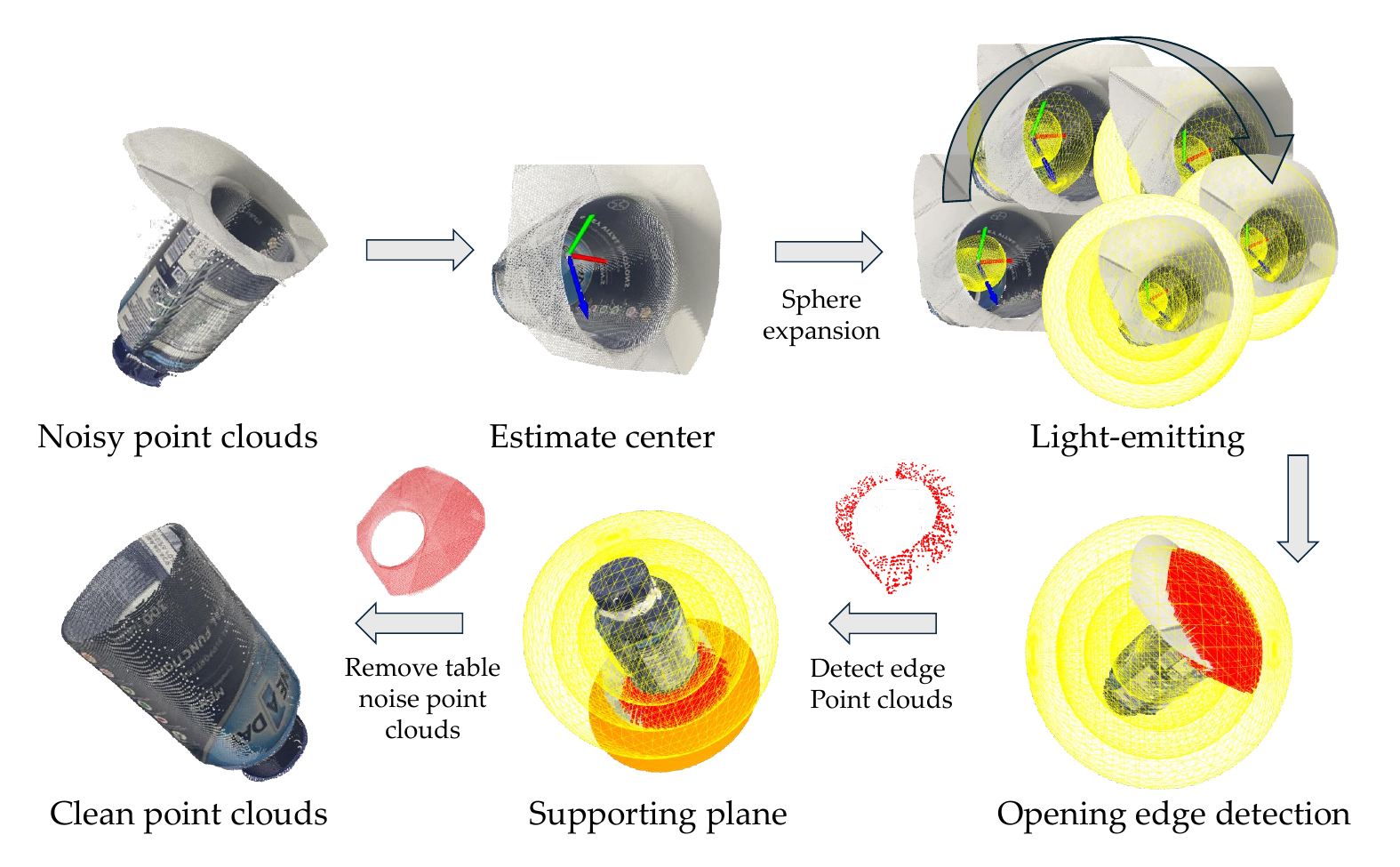}
    \caption{\textbf{Supporting-plane noise removal mechanism.} A virtual light sphere expands from the object center to identify openings points, enabling filtering of table noise.}
    \vspace{-0.3cm}
    \label{fig:denoise-pipeline}
\end{figure}

\noindent\textbf{Point Clouds Denoising.}  
Detecting openings or cavities in 3D point clouds is a fundamental step in shape understanding for denoising table noise. 
We propose a purely geometric method that progressively expands a virtual light sphere from the object's center and tracks uncovered regions on the spherical sampling space. 
The algorithm automatically detects openings and extracts boundary rim points without manual boundary selection, as illustrated in Fig.~\ref{fig:denoise-pipeline}.

Given a point cloud $\mathcal{P}$, we first estimate a geometric center
$\mathbf{c} = \mathrm{mean}(\mathcal{P})$.
We then discretize the unit sphere into $F$ directions
$\{\mathbf{d}_i\}_{i=1}^{F}$ 
using a Fibonacci spiral distribution~\cite{gonzalez2010measurement}, forming a uniform sampling domain $\mathcal{D}$. Each point $\mathbf{p}_i$ defines a normalized direction
$\hat{\mathbf{v}}_i = (\mathbf{p}_i - \mathbf{c}) / \|\mathbf{p}_i - \mathbf{c}\|$
and is assigned to its nearest angular bucket direction $\mathbf{d}_j$
if $\hat{\mathbf{v}}_i \cdot \mathbf{d}_j \ge \cos(\theta_{\text{tolerance}})$.
We then iteratively expand a virtual sphere centered at $\mathbf{c}$ with radius $r_t$, along the sampled directions.
\begin{equation}
r_{t+1} = r_t + \Delta r, \quad
r_{\min}(j) = \min_{i \in \text{bucket } j} \|\mathbf{p}_i - \mathbf{c}\|.
\label{eq:sphere}
\end{equation}
A direction $\mathbf{d}_j$ is marked as \textit{hit} once any assigned point enters the sphere,
and unhit directions form a binary mask $\mathcal{U}_t$. 
The largest unhit connected component is identified as an opening or cavity after iterations.
We denote by $\mathcal{N}(j)=\{k\mid (j,k)\in \mathcal{P}\}$ the neighborhood of bucket $j$.
Boundary buckets are defined as unhit directions adjacent to hit ones:
\begin{equation}
\mathcal{B} = \{ j \in \mathcal{U} \mid \exists\, k \in \mathcal{N}(j),\, \text{hit}(k) = 1 \}.
\label{eq:boundary}
\end{equation}
For each boundary $\mathbf{d}_j$, the farthest point within tolerance is chosen as a rim sample $\mathbf{p}^*_j$.
Connected components on the spherical adjacency graph are extracted to identify large uncovered regions.
Let $\{\mathcal{C}_m\}$ denote all connected components of the unhit set $\mathcal{U}_t$, $\mathcal{C}_{\max}$ is the largest connected region.
The principal opening direction is then computed by averaging the largest unhit component:
$
\mathbf{n}_{\text{open}} =
\frac{\sum_{j \in \mathcal{C}_{\max}} \mathbf{d}_j}
{\|\sum_{j \in \mathcal{C}_{\max}} \mathbf{d}_j\|}.
\label{eq:plane}
$
Finally, we fit a plane to rim points $\{\mathbf{p}^*_j\}$ via SVD to remove the supporting surface noise points.
Compared with RANSAC plane fitting~\cite{fischler1981random}, which can only segment dominant planes, our method can detect cavity openings, enabling the identification of ring-shaped planes surrounding the opening.

\noindent\textbf{Real-World Scale Alignment.}  
Since VGGT produces point clouds that are not in a world-scale metric, 
we estimate a global scale factor to align the reconstructions with world coordinates. 
Even with known intrinsic, according to the pinhole camera model, 
monocular geometry cannot determine absolute scale. 
Therefore, we calibrate the scale using a marker of known physical dimensions within the scene. 

\noindent\textbf{Mesh Simplification.}  
To maintain real-time performance in the digital twin, we apply an adaptive mesh decimation that reduces vertex count while preserving geometric fidelity.  
To avoid over-smoothing across sharp edges, we use an angle-based gating weight  
$w_{ij}=1$ if $\theta_{ij}\le\theta_{\mathrm{th}}$ and $0$ otherwise,  
where $\theta_{\mathrm{th}}$ is a feature threshold (e.g., $30^\circ$) and  
$\theta_{ij} = \arccos(\mathbf{n}_i^\top \mathbf{n}_j)$ for all $j \in \mathcal{N}(i)$. 
Compared to uniform mesh decimation, this feature-aware smoothing preserves sharp edges around handles and object rims, which are critical for accurate grasp planning.
All processed point clouds, meshes, and their id are stored in a memory bank, which serves for scene synchronization in Stage~II.

\subsection{Online Digital Twin Synchronization}
The second stage of our system focuses on real-time object tracking and safe grasp execution through continuous perception, planning synchronization between the real and digital environments. This stage consists of object-state estimation, geometry updates, grasp generation, and reactive planning. Stage~II accepts either human clicked prompts or object detections for masks generation. 

\begin{figure}[t]
    \centering
    \includegraphics[width=1\linewidth]{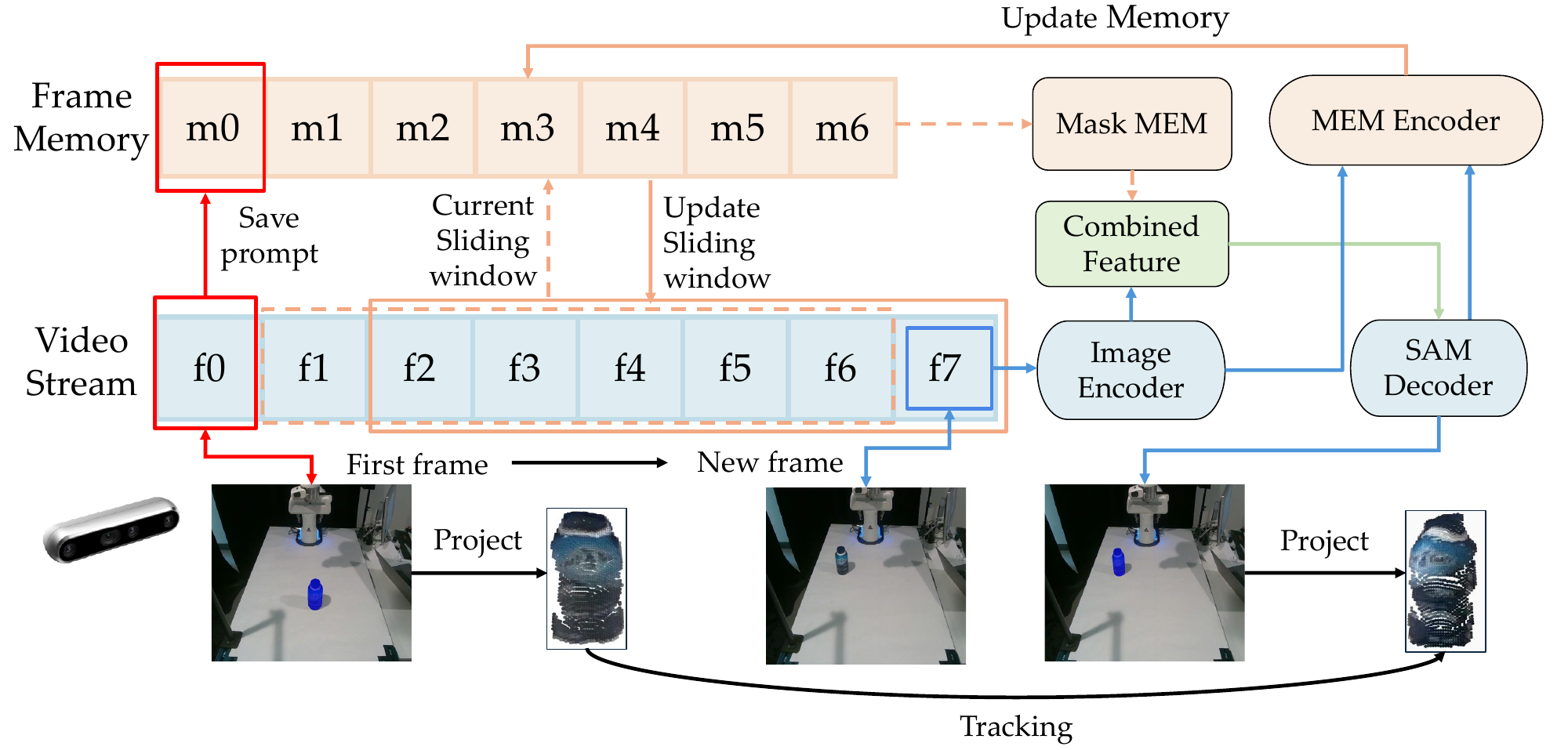}
    \caption{
    \textbf{Overview of the camera predictor module.} 
    The red solid line indicates that the first frame is persistently stored in the frame memory.
    The yellow dashed line represents saving frames from the previous time step into memory.
    The blue and green solid lines denote the data flow and processing of the current frame.
    Together, the sliding-window mechanism enables real-time video segmentation and object-level point cloud tracking with temporally consistent memory updates.
    }
    \vspace{-0.3cm}
    \label{fig:sam2-realtime}
\end{figure}

\noindent\textbf{Real-time Point Clouds Segmentation.}
Unlike standard video-mask propagation pipelines~\cite{ravi2025sam} that are primarily designed for offline videos, our module is implemented for live RGB-D camera streams.
As shown in Fig.~\ref{fig:sam2-realtime}, our real-time segmentation module conducts continuous inference on incoming frames to output masks $\mathcal{S}_p$, which are then projected onto the full point cloud to yield partial object point clouds $\mathcal{X}_p$.
To support streaming input, we design a sliding-window inference mechanism that maintains temporal consistency of object masks across frames, enabling continuous 3D segmentation and tracking under occlusions. At image size 640×480, our method runs at 15 Hz on RTX 4090.

\noindent\textbf{Object Pose Estimation and Tracking.}
 We select the tracking backend by GPU memory and asset availability. Our real-time point-cloud segmentation, combined with colored-ICP~\cite{park2017colored} for known assets, provides a lower-memory option while retaining point-cloud tracking for unseen objects. Alternatively, FoundationPose~\cite{wen2024foundationpose} requires more GPU memory. 
Their trade-offs and suitable operating conditions will be evaluated in Sec.~\ref{sec:backend_comparison}.

\noindent\textbf{Grasp Pose Generation and Obstacle Representation.}
For unseen target objects, we directly apply GraspGen to the partial point cloud. 
For registered seen objects, we replace partial observation with a complete  memory-bank point cloud $\mathcal{X}_m$ to predict grasp poses 
$\{\mathbf{T_{gripper}} = f_{\text{GraspGen}}(\mathcal{X}_m)\}$. 
This replacement mitigates the uncertainty from occlusion and single-view perception, yielding more stable and accurate grasp pose estimation.
For unseen obstacles, we dynamically generate multi-convex hulls~\mbox{\cite{wei2022approximate}} from the segmented point cloud for collision modeling.
For known obstacles, aligned object meshes and poses are directly imported into the digital twin for real-time collision checking.

\noindent\textbf{Out-of-View Recovery.}
After an object re-enters the camera view, both tracking backends perform re-detection, re-segmentation, and object-ID reassociation. Our real-time point-cloud tracking and ICP branch then resumes tracking, whereas the FoundationPose branch additionally requires mesh re-registration before pose tracking, resulting in longer recovery latency and additional GPU memory overhead.

\noindent\textbf{Motion Planning with Real2sim2real Synchronization.}
Finally, motion planning and control are performed using cuRobo's MPC framework. At each control step, the robot’s joint states are synchronized with the digital twin, where cuRobo computes an optimized short-horizon trajectory under real-time collision constraints. Only the first control action $\{\mathbf{a}_0\}$ from the predicted trajectory $\mathbf{A}_{t:t+H}=\{\mathbf{a}_0,\mathbf{a}_1,\ldots,\mathbf{a}_H\}$ is executed on the real robot, followed by continuous replanning with environment updates, achieving closed-loop synchronization between simulation and reality.

\begin{figure}[!t]
    \centering
    \includegraphics[width=1\linewidth]{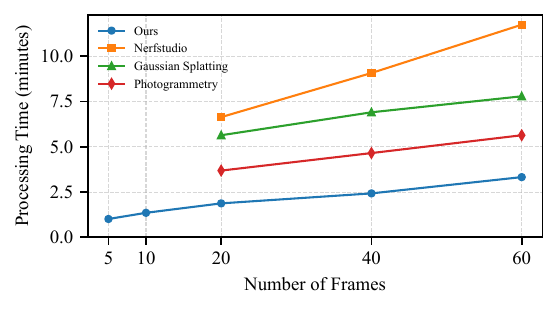}
    \caption{
    \textbf{Comparison of processing time given different numbers of input images.}
    The reported time includes both reconstruction and segmentation.
    In practice, our method achieves reliable reconstruction with as few as 5-10 input images.
    Results for 5 and 10 images are not reported for the baselines, as they fail to estimate camera extrinsics.}
    \vspace{-0.3cm}
    \label{fig:recon_line}
\end{figure}

\section{Experiments}
\label{sec:experiments}
We evaluate SyncTwin across four dimensions: 
(1) the efficiency of the fast 3D reconstruction pipeline;
(2) the pose accuracy of the synchronized digital twin under both unoccluded and occluded conditions;
(3) obstacle avoidance performance under dynamic and single-view occluded conditions; and 
(4) grasp success rate under single-view occlusion. 

\subsection{Experiment Setup}

Experiments use a Franka Panda arm, an Intel RealSense D455 RGB-D camera mounted above and in front of the workspace, and an iPhone 12 for RGB image capture.
Input point clouds are voxel-downsampled at 3~mm.
Isaac Sim~4.0 and cuRobo MPC run on a single NVIDIA RTX~4090 GPU, with scene synchronization asynchronous to the 10~Hz planning loop. Robot velocity scaling is set to 0.2; test objects include bottles, cans, cups, and boxes.

\subsection{Baselines and Metrics}
\textbf{Baselines.}  
(1) For \textit{3D reconstruction}: we compare against Photogrammetry~\cite{schonberger2016structure}, NeRF ~\cite{tancik2023nerfstudio}, and Gaussian Splatting~\cite{kerbl20233d}, which are common choices for digital twin construction in related systems; (2) For \textit{obstacle avoidance}: we adopt NVBlox~\cite{millane2024nvblox} as the baseline voxel-mapping method.

\textbf{Ablation Studies.}
We analyze performance across the following settings:
(1)\textit{Tracking and Recovery}: comparing colored-ICP and FoundationPose in object tracking and out-of-view recovery.
(2) \textit{Mask Expansion and Denoising}: using variants without mask expansion and without denoising as baselines, evaluating their
quality against our full mechanism;
(3) \textit{Object Completion}: 
comparing grasp poses predicted from single-view partial point clouds (baseline) versus those from complete geometry produced by SyncTwin (ours).

\noindent\textbf{Evaluation Metrics.}  
Depending on the experiment, we report the following metrics:  
reconstruction time (min), dependency on the number of input images \(N_{\min}\), pose accuracy (\%), obstacle avoidance rate (\%), and grasp success rate (\%). 

Pose accuracy is evaluated using normalized translation and rotation accuracy scores.
For translation, let $\Delta \mathbf{t}_{\mathrm{real}}^{(i)}$ and $\Delta \mathbf{t}_{\mathrm{twin}}^{(i)}$ denote the object translations in the real world and in the digital twin for the $i$-th trial.
The translation error is defined as
$e_{\mathrm{trans}}^{(i)} =
\left\|
\Delta \mathbf{t}_{\mathrm{real}}^{(i)} -
\Delta \mathbf{t}_{\mathrm{twin}}^{(i)}
\right\|_2 ,$
and the translation accuracy is computed as
$\mathrm{Acc}_{\mathrm{trans}} =
1 - \frac{e_{\mathrm{trans}}}{t_{\mathrm{real}}}$.
For rotation, object orientations are represented as unit quaternions.
Given the real and simulated orientations $\mathbf{q}_{\mathrm{real}}^{(i)}$ and $\mathbf{q}_{\mathrm{twin}}^{(i)}$, the angular rotation error is computed as
$e_{\mathrm{rot}}^{(i)} =
2 \arccos\!\left(\left|q_{\mathrm{rel}}^{w}\right|\right)$,
where $\mathbf{q}_{\mathrm{rel}}^{(i)} =
\left(\mathbf{q}_{\mathrm{real}}^{(i)}\right)^{-1}
\otimes
\mathbf{q}_{\mathrm{twin}}^{(i)}$.
The rotation accuracy is defined as
$\mathrm{Acc}_{\mathrm{rot}} =
1 - \frac{e_{\mathrm{rot}}}{r_{\mathrm{real}}}$.
Higher accuracy values indicate better alignment between the digital twin and the real world.

\label{sec:metrics}

For avoidance tests, we define 3-level outcome levels: FA (full avoidance, no contact with the object, 1.0), 
EA (edge avoidance, slight contact without displacing the object, 0.8), 
and CO (collision, noticeable contact that significantly moves the object, 0.0). Weighted success rate:
$\text{SR}=\frac{N_{\mathrm{FA}}+0.8\,N_{\mathrm{EA}}+0.0\,N_{\mathrm{CO}}}{N}\times 100\%$.

\begin{figure}[!t]
    \centering
    \includegraphics[width=1\linewidth]{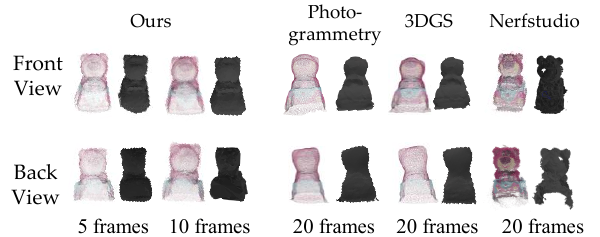}
   \caption{
    \textbf{Reconstruction results with different numbers of input images.}
    In each column, the left figure shows the point clouds, and the right one shows the untextured mesh.
    }
    \vspace{-0.3cm}
    \label{fig:recon_comparison}
\end{figure}

\begin{table*}[!t]
\centering
    \caption{\textbf{Comparison of obstacle-avoidance performance between NVBlox and SyncTwin in dynamic environments.}}
    \label{tab:exp2}
    
\setlength{\tabcolsep}{1.5pt}
\begingroup
\fontsize{6.5pt}{8.5pt}\selectfont
\renewcommand{\arraystretch}{0.9}
\setlength{\aboverulesep}{1pt}
\setlength{\belowrulesep}{1pt}
\begin{tabular*}{\textwidth}{@{\extracolsep{\fill}}llc*{26}{c}}
\toprule

\multirow{3}{*}{Method} &
\multirow{3}{*}{Motion} &
\multirow{3}{*}{\shortstack{Trans. Velocity}} &
\multicolumn{13}{c}{\textbf{Unseen}} &
\multicolumn{13}{c}{\textbf{Seen (Memory Bank)}} \\

\cmidrule(lr){4-16}
\cmidrule(lr){17-29}

& & &
\multicolumn{3}{c}{Box1} &
\multicolumn{3}{c}{Box2} &
\multicolumn{3}{c}{Box3} &
\multicolumn{3}{c}{Box4} &
$SR_{\mathrm{unseen}}(\%)$ &
\multicolumn{3}{c}{Box1*} &
\multicolumn{3}{c}{Box2*} &
\multicolumn{3}{c}{Box3*} &
\multicolumn{3}{c}{Box4*} &
$SR_{\mathrm{seen}}(\%)$ \\

\cmidrule(lr){4-6}
\cmidrule(lr){7-9}
\cmidrule(lr){10-12}
\cmidrule(lr){13-15}
\cmidrule(lr){17-19}
\cmidrule(lr){20-22}
\cmidrule(lr){23-25}
\cmidrule(lr){26-28}

& & &
FA & EA & CO &
FA & EA & CO &
FA & EA & CO &
FA & EA & CO &
mean $\pm$ SD &
FA & EA & CO &
FA & EA & CO &
FA & EA & CO &
FA & EA & CO &
mean $\pm$ SD \\

\midrule

\multirow{2}{*}{\textbf{NVBlox}}
& SelfRot
& 0 m/s
& 6 & 4 & 10
& 9 & 5 & 6
& 3 & 4 & 13
& 3 & 11 & 6
& 50.3 $\pm$ 15.1
&
-- & -- & --
& -- & -- & --
& -- & -- & --
& -- & -- & --
& -- \\

& EnterTraj
& 0.2 m/s
& 3 & 9 & 8
& 4 & 9 & 7
& 1 & 6 & 13
& 0 & 3 & 17
& 37.0 $\pm$ 20.4
&
-- & -- & --
& -- & -- & --
& -- & -- & --
& -- & -- & --
& -- \\

\midrule

\multirow{3}{*}{\textbf{SyncTwin}}
& SelfRot
& 0 m/s
& 12 & 7 & 1
& 14 & 5 & 1
& 11 & 7 & 2
& 9 & 9 & 2
& \textbf{85.5 $\pm$ 4.2}
&
18 & 1 & 1
& 16 & 4 & 0
& 15 & 5 & 0
& 13 & 6 & 1
& \textbf{93.5 $\pm$ 3.1} \\

& \multirow{2}{*}{EnterTraj}
& 0.2 m/s
& 8 & 8 & 4
& 11 & 6 & 3
& 8 & 7 & 5
& 7 & 8 & 5
& \textbf{71.5 $\pm$ 5.4}
&
12 & 7 & 1
& 12 & 5 & 3
& 10 & 6 & 4
& 9 & 7 & 4
& \textbf{78.8 $\pm$ 6.9} \\

& & 0.5 m/s
& 4 & 9 & 7
& 1 & 9 & 10
& 2 & 7 & 11
& 0 & 2 & 18
& \textbf{35.8 $\pm$ 20.1} 
& 6 & 8 & 6
& 1 & 10 & 9
& 3 & 7 & 10
& 0 & 7 & 13
& \textbf{44.5 $\pm$ 13.9} \\

\bottomrule
\end{tabular*}\par
\vspace{2pt}
{\scriptsize\parbox{\textwidth}{Unseen/Seen* denotes objects without/with memory-bank assets. Trans. Velocity is obstacle translational speed (m/s). SR is the mean $\pm$ sample SD of weighted scores (Sec.~\ref{sec:metrics}) across four boxes, with 20 trials per box per condition.}}
\endgroup
\vspace{-0.3cm}
\end{table*}

\subsection{Fast 3D Reconstruction}
We evaluate the efficiency and input-image number dependency of SyncTwin’s 3D reconstruction module, comparing against Photogrammetry, 3DGS, and Nerfstudio.
Each method reconstructs the same object using 5, 10, 20, 40, and 60 RGB images as input, and all approaches obtain object assets through multi-view projection-based segmentation.

\noindent\textbf{Reconstruction Time.}
As shown in Fig.~\ref{fig:recon_line}, SyncTwin achieves the shortest reconstruction time across all settings.
It generates a simulation-ready mesh in approximately 1-2 minutes using only 5-10 input images, whereas Photogrammetry, 3DGS, and Nerfstudio require more input images and substantially longer processing times, typically ranging from 4 to 8 minutes.
As a result, SyncTwin enables significantly faster digital twin construction in new real-world scenarios.

\noindent\textbf{Dependency on the Number of Input Images.}
SyncTwin requires only 5--10 RGB images to generate usable meshes, whereas competing methods typically need over 20 to avoid failures from unstable camera extrinsic optimization. This reduces reconstruction cost and deployment overhead. Since the runtime in Fig.~\ref{fig:recon_line} excludes image acquisition and data transfer, fewer images also save collection and transmission time, enabling efficient construction under limited viewpoints.

\begin{table*}[t]
\centering
\setlength{\tabcolsep}{3pt}
\caption{\textbf{Pose accuracy and four-object tracking efficiency of SyncTwin.}}
\label{tab:pose_accuracy}
\begingroup
\fontsize{6.5pt}{8.5pt}\selectfont
\renewcommand{\arraystretch}{0.9}
\setlength{\aboverulesep}{1pt}
\setlength{\belowrulesep}{1pt}
\begin{tabular*}{\textwidth}{@{\extracolsep{\fill}}ll*{12}{c}}
\toprule
 & & \multicolumn{2}{c}{Box} & \multicolumn{2}{c}{Cup} & \multicolumn{2}{c}{Bottle} & \multicolumn{2}{c}{Ketchup} & \multicolumn{2}{c}{Overall (mean $\pm$ SD)} & \multicolumn{2}{c}{Four-object tracking} \\
\cmidrule(lr){3-4} \cmidrule(lr){5-6} \cmidrule(lr){7-8} \cmidrule(lr){9-10}
\cmidrule(lr){11-12} \cmidrule(lr){13-14}
Method & Condition
& Trans. & Rot.
& Trans. & Rot.
& Trans. & Rot.
& Trans. & Rot.
& Trans. & Rot.
& Rate (FPS) & GPU memory (GiB) \\
\midrule
\multirow{2}{*}{RT-ICP} & Unoccluded
& 95.1 & 78.3
& 96.3 & 88.7
& 94.8 & 90.4
& 94.7 & 82.0
& 95.23 $\pm$ 0.74 & 84.85 $\pm$ 5.68
& 3.43 & \textbf{5.70 $\pm$ 0.50 / 6.10} \\
 & Occluded
& 85.2 & 70.4
& 87.7 & 78.2
& 90.3 & 82.6
& 91.5 & 76.3
& 88.68 $\pm$ 2.81 & 76.88 $\pm$ 5.06
& -- & \-- \\
\midrule
\multirow{2}{*}{FP} & Unoccluded
& 97.1 & 93.2
& 97.4 & 92.1
& 98.7 & 95.3
& 97.4 & 96.8
& \textbf{97.65 $\pm$ 0.71} & \textbf{94.35 $\pm$ 2.10}  
& \textbf{11.57} & 7.70 $\pm$ 0.38 / 16.03 \\
 & Occluded
& 91.5 & 89.3
& 94.2 & 90.8
& 96.3 & 92.6
& 95.1 & 93.7
& \textbf{94.28 $\pm$ 2.04} & \textbf{91.60 $\pm$ 1.94}
& -- & -- \\
\bottomrule
\end{tabular*}\par
\vspace{2pt}
{\scriptsize\parbox{\textwidth}{RT-ICP: real-time tracking + colored-ICP; FP: FoundationPose. Translation (Trans.) and rotation (Rot.) accuracy are reported in \%.
Overall accuracy is the mean $\pm$ sample SD across the four objects. The rightmost columns report FPS and GPU memory (mean $\pm$ SD/peak) for simultaneous four-object tracking under unoccluded conditions.}}
\endgroup
\vspace{-0.3cm}
\end{table*}

Furthermore, the qualitative comparison in Fig.~\ref{fig:recon_comparison} shows that our method preserves fine-grained geometric details even with very few input images, as our mesh simplification algorithm effectively maintains high geometric fidelity while reducing the number of mesh vertices, enabling faster simulation performance. For instance, the \emph{bear's ear shape} remains well preserved with only 5-10 frames, demonstrating both low image dependency and strong geometric consistency. 

These properties make SyncTwin a state-of-the-art solution for fast 3D geometry asset generation: the system can construct high-quality, simulation-ready object meshes in about 1 minute, enabling rapid reconstruction of object assets. In our setup, the complete procedure from image acquisition to simulation-ready assets takes less than 5 minutes, including approximately 2 minutes for image capture/upload, 5s for prompt initialization and masks generation, 1 min for automatic reconstruction and asset processing, and about 1 minute for the operator to launch the required scripts.

\subsection{Obstacle Avoidance under Occlusion}

We adopt cuRobo's built-in obstacle avoidance benchmark to evaluate the avoidance success rate, where the robot repeatedly moves between two target points.

We test NVBlox and SyncTwin on unseen (not stored in the memory bank) obstacles. SyncTwin uses the real-time point-cloud tracking and colored-ICP configuration (RT-ICP). Additionally, SyncTwin is further evaluated on seen (stored in memory bank) objects to study how prior geometry improves safety under single-view occlusion.
Two motion patterns are considered: \textbf{SelfRot} for in-place rotation and \textbf{EnterTraj} for trajectory-blocking obstacle motion.
Each condition is repeated for 20 trials, in total $N = 20 \times 4 = 80$. 
We test four boxes with sizes of 
Box1: $10{\times}10{\times}10$~cm, 
Box2: $20{\times}20{\times}20$~cm, 
Box3: $10{\times}20{\times}30$~cm, and 
Box4: $20{\times}22{\times}35$~cm as obstacles.
Based on these experimental settings, the results are summarized as follows:

\noindent\textbf{Unseen Object Performance.}  
As shown in Table~\ref{tab:exp2}, SyncTwin consistently outperforms NVBlox when encountering unseen obstacles under single-view dynamic occlusion. NVBlox exhibits failure modes such as: (1) unstable voxelization for small objects (e.g., Box1) due to sparse depth returns; (2) misclassification of limited height objects (e.g., Box3) as part of the tabletop caused by inconsistent depth estimation; and (3) trajectory intersections when large obstacles (e.g., Box4) fall outside the sensor’s visible region.  
Across all these cases, SyncTwin maintains markedly higher obstacle-avoidance success rates, demonstrating stronger robustness to object scale, occlusion, and limited viewpoint.

\noindent\textbf{Seen Object Performance.}  
Table~\ref{tab:exp2} shows further gains for objects stored in the memory bank. Complete geometry from Stage~I reduces failures involving small or thin objects, highlighting the value of geometric priors for reliable obstacle avoidance under partial observability.

Table~\ref{tab:exp2} also shows that increasing obstacle speed from 0.2 to 0.5~m/s degrades SyncTwin's avoidance performance for both seen and unseen objects, highlighting its limited responsiveness to faster scene dynamics.
At 0.5~m/s, collisions are predominantly caused by perception latency, the planner's lack of obstacle-motion prediction, and execution lag.

\subsection{Ablation Studies}
\phantomsection
\label{sec:backend_comparison}
\noindent\textbf{Tracking and Recovery.}
We compare real-time point-cloud tracking with colored-ICP registration (RT-ICP) and FoundationPose on four objects (box, cup, bottle, and ketchup bottle) under unoccluded and occluded conditions, using the same Stage~I assets and RGB-D observations.
Each object is translated by 15~cm along the Cartesian axes and rotated by 90 degrees around the roll, pitch, and yaw axes, with each motion repeated 50 times.
Table~\ref{tab:pose_accuracy} shows that FoundationPose achieves higher pose accuracy across objects and occlusion conditions, and faster simultaneous four-object tracking. RT-ICP offers lower accuracy and speed but requires less GPU memory. Table~\ref{tab:sync_latency} confirms the trade-off between lower steady-state latency and higher memory usage for FoundationPose. In our configuration, FoundationPose requires a reference asset, whereas RT-ICP retains a point-cloud branch for completely unseen objects without stored assets. For these unseen objects, costly online V-HACD generation limits synchronization to about 5~Hz. When all objects have stored assets, their collision geometry can be reused, enabling faster synchronization.

Both backends use re-detection, re-segmentation, and object-ID
reassociation for out-of-view recovery. The FoundationPose branch additionally requires mesh re-registration, increasing recovery
latency and peak GPU memory (Table~\ref{tab:sync_latency}). Overall, FoundationPose is preferable when accuracy and tracking speed are priorities and assets and GPU memory are available; RT-ICP is better suited to memory-constrained deployment, asset-free object handling, and faster recovery.

\begin{table}[t]
\centering
\caption{\textbf{Runtime breakdown and GPU memory comparison.}}

\label{tab:sync_latency}

\begingroup
\scriptsize
\renewcommand{\arraystretch}{0.95}
\setlength{\tabcolsep}{3pt}

\begin{tabularx}{\linewidth}{
    @{}
    >{\raggedright\arraybackslash}X
    >{\centering\arraybackslash}p{0.27\linewidth}
    >{\centering\arraybackslash}p{0.27\linewidth}
    @{}
}
\toprule
\textbf{Component} & \textbf{RT-ICP} (mean $\pm$ SD)  & \textbf{FP} (mean $\pm$ SD)  \\
\midrule

Camera preprocessing
& 1.35 $\pm$ 0.71 ms
& 1.28 $\pm$ 0.41 ms \\

Detection$^{*}$
& 5.83 $\pm$ 0.55 ms
& 5.90 $\pm$ 0.79 ms \\

Segmentation$^{*}$
& 134.36 $\pm$ 40.48 ms
& 138.51 $\pm$ 44.99 ms \\

FP object registration$^{*}$
& N/R
& 650.13 $\pm$ 254.20 ms \\

Point-cloud extraction
& 1.78 $\pm$ 0.52 ms
& N/R \\

V-HACD generation$^{+}$
& 111.41 $\pm$ 20.18 ms
& N/R \\

Tracking
& 63.34 $\pm$ 9.55 ms
& 9.97 $\pm$  2.45 ms \\

Pose estimation
& 7.15 $\pm$ 1.10 ms
& 11.55 $\pm$ 2.47 ms \\

Digital twin scene update
& 0.31 $\pm$ 0.03 ms
& 0.29 $\pm$ 0.04 ms \\

cuRobo planning latency$^{\bullet}$
& 49.31 $\pm$ 0.47 ms
& 53.80 $\pm$ 0.58 ms \\

\midrule
\multicolumn{3}{@{}l}{\textit{System-level metrics}} \\

O2U time/rate (ms/Hz) 
& 73.93 $\pm$ 9.74 / 13.53 
& \textbf{25.09 $\pm$ 2.87 / 39.86}  \\

O2U time/rate$^{+}$ (ms/Hz)
& \textbf{185.34 $\pm$ 21.67 / 5.40}
& N/R \\

Initialization/Recovery time$^{*}$ (ms) 
& \textbf{214.12  $\pm$ 46.08}  
& 819.63 $\pm$ 46.35  \\

GPU memory (GiB, mean/peak)
& \textbf{5.59 $\pm$ 0.51 / 5.99} 
& 7.56 $\pm$ 0.37 / 15.86  \\

GPU memory$^{*}$ (GiB, mean/peak)
& \textbf{6.23 $\pm$ 0.23 / 6.69}
& 9.34 $\pm$ 2.58 / 17.08  \\

\bottomrule
\end{tabularx}

\vspace{2pt}
\parbox{\linewidth}{RT-ICP: real-time point clouds tracking (ours) + colored-ICP; FP: FoundationPose. O2U denotes observation-to-update latency.
Unmarked components constitute the steady-state O2U time for one seen object.
$^{+}$ denotes components additionally required for completely unseen objects,
and $^{*}$ denotes components additionally required during initialization/recovery.
Thus, O2U$^{+}$ is the O2U time plus all $^{+}$-marked components for one seen and one completely unseen object, while
initialization/recovery time is the O2U time plus all $^{*}$-marked components. $^{\bullet}$ denotes asynchronous motion planning, which does not block scene updates and is excluded from O2U latency. N/R: not required.}
\endgroup
\vspace{-0.3cm}
\end{table}

\begin{figure}[!t]
    \centering
    \includegraphics[width=1\linewidth]{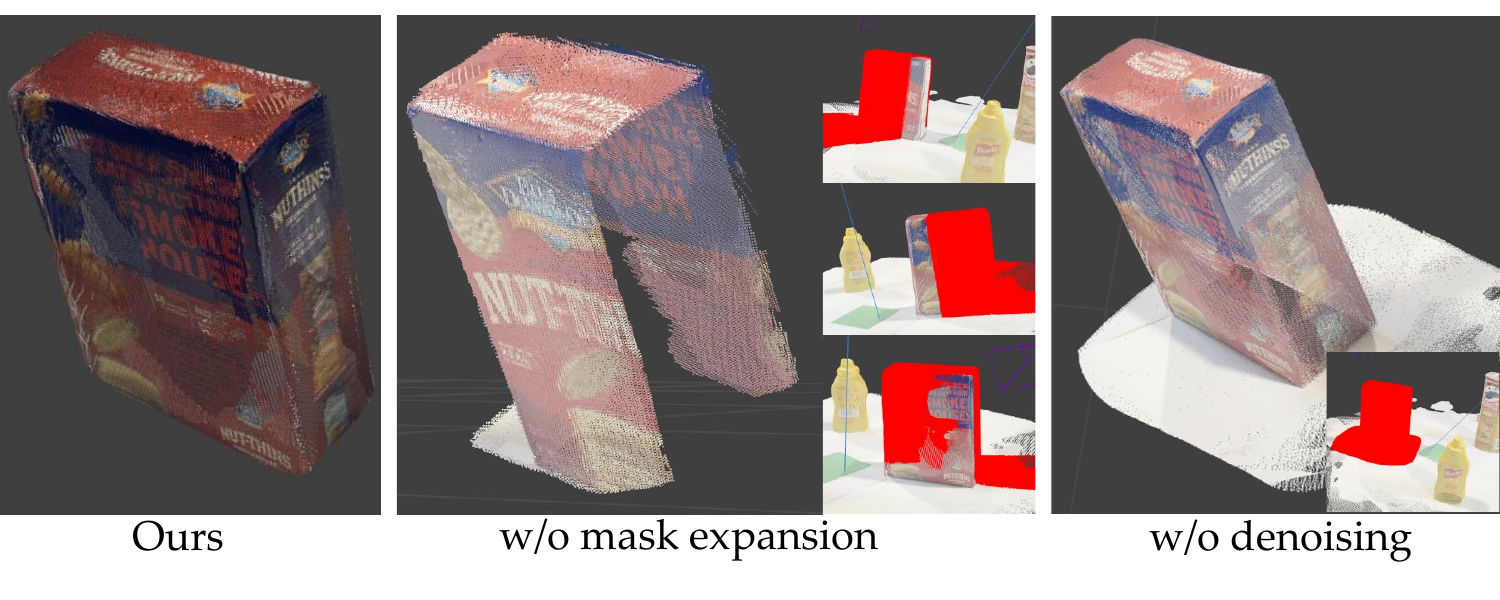}
     \vspace{-0.8cm}
   \caption{
   \textbf{Comparison of segmentation-based 3D reconstruction results.} Left: Our method generates a complete and clean object mesh without support-plane noise. Middle: w/o mask expansion, multi-view projections (red) fail to fully overlap, and their intersection leads to missing object geometry. 
    Right: w/o denoising, the intersection of projected regions incorrectly preserves support-plane points, introducing noise into the reconstructed point cloud.}
    \label{fig:ablation1}
    \vspace{-0.4cm}
\end{figure}

\noindent\textbf{Mask Expansion and Denoising.}
We ablate two variants: \textit{w/o mask expansion} and \textit{w/o denoising}.
As shown in Fig.~\ref{fig:ablation1}, mask expansion removal causes inconsistent cross-view projections and incomplete object reconstruction, while disabling support-plane denoising yields noisy meshes.
The results verify that mask expansion and support-plane denoising are indispensable for high-quality 3D asset reconstruction.

\begin{figure}[t]
    \centering
    \includegraphics[width=1\linewidth]{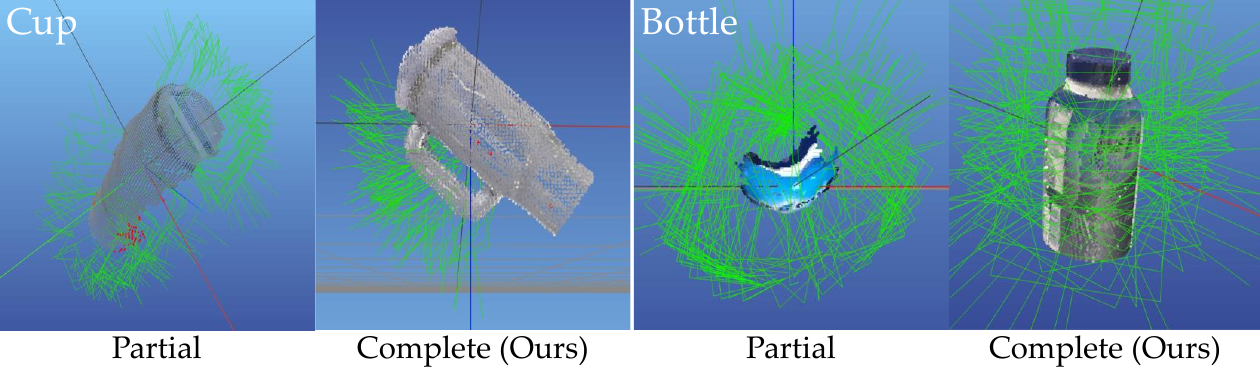}
    \vspace{-0.6cm}
    \caption{
    \textbf{Comparison of grasp candidates generated from Ours (w/o Completion) and Ours.}
    The example objects include a handled cup (left pairs) and a bottle (right pairs). 
    Green lines visualize predicted gripper poses.
    }
    \label{fig:ablation2}
    \vspace{-0.5cm}
\end{figure}

\noindent\textbf{Object Completion.}
We evaluate the effect of object geometric completeness on grasp generation.
When grasp poses are generated from single-view partial point clouds, the limited geometry often leads to incomplete or incorrect grasp configurations. For example, as shown in Fig.~\ref{fig:ablation2}, the handle of the cup is not perceived, resulting in unsafe or colliding grasps. 
In contrast, after SyncTwin aligns the object with its complete mesh from the asset library, the grasp generator produces denser, more accurate, and geometrically feasible grasps. 

\subsection{Discussions}
Our work provides a segmentation approach that covers most objects, while some challenging cases remain when applied to learning-based reconstructed point clouds.
For instance, the supporting-plane denoising strategy may fail for tall or weakly supported objects such as wine glasses.
This work makes an early attempt at tackling such challenging cases, with further open problems remaining.
This work emphasizes a principled view of digital twins as real-time synchronization of environmental geometry, rather than merely visual or appearance-level simulation.
Although several prior works are referred to as digital twins, they primarily focus on visual consistency and perform a single offline scene reconstruction, without explicitly addressing continuous state synchronization between the real world and simulation.
As a result, such systems are fundamentally closer to visual simulation than to a digital twin with real-time synchronization capability.
By contrast, SyncTwin continuously tracks and synchronizes object geometry and pose to capture environmental dynamics and maintain reliable real-to-sim-to-real consistency.
However, the performance in dynamic scenes is also limited by our cuRobo configuration. It plans using scene representations without predicting obstacle trajectories, resulting in worse obstacle avoidance as obstacle speed increases.
Abrupt obstacle accelerations or sudden direction changes, which we did not explicitly evaluate, would further challenge this reactive loop, since such changes are captured only at the next scene update and acted upon at the next replanning cycle, during which the robot follows a trajectory planned for the outdated obstacle state.
Moreover, the proposed digital twin synchronization also provides a foundation for safe real-robot reinforcement learning.
By synchronizing real-world geometry into simulation, the system enables collision-aware training and execution.
This capability is particularly important for post-training and fine-tuning learning-based policies on physical robots.

\section{Conclusion}

We introduced SyncTwin, a digital twin framework that unifies fast RGB-only 3D reconstruction with real-time scene synchronization for safety-aware and robust execution in dynamic, partially occluded environments.
Future work includes merging Stage~I construction into Stage~II synchronization, extending SyncTwin to appearance-level consistency for realistic visuomotor policy training~\cite{yan2026matters}, and adopting distributed/multi-GPU designs to decouple perception, synchronization, and planning for higher dynamic-scene responsiveness. We will also explore probabilistic obstacle-state modeling for uncertainty-aware planning. These directions will enhance digital twin capabilities.

\bibliographystyle{IEEEtran}
\bibliography{references}

\end{document}